\pdfoutput=1

\documentclass[11pt]{article}

\usepackage{naacl2021}

\usepackage{times}
\usepackage{latexsym}
\usepackage{graphicx}
\usepackage{subcaption}
\usepackage{booktabs}
\usepackage{amsmath}
\usepackage{lipsum}
\usepackage{color,soul}

\usepackage[T1]{fontenc}

\usepackage[utf8]{inputenc}

\usepackage{microtype}

\title{Hurdles to Progress in Long-form Question Answering}

\author{Kalpesh Krishna$^{\spadesuit\,*}$ \And Aurko Roy$^\diamondsuit$ \\\\ $^\spadesuit$University of Massachusetts Amherst, $^\diamondsuit$Google Research \\ \texttt{\{kalpesh,miyyer\}@cs.umass.edu}\\ \texttt{aurkor@google.com} \And Mohit Iyyer$^\spadesuit$}

 \newcommand{\namedref}[2]{\hyperref[#2]{#1~\ref*{#2}}}

\newcommand{\sectionref}[1]{\namedref{Section}{#1}}
\newcommand{\tableref}[1]{\namedref{Table}{#1}}
\newcommand{\figureref}[1]{\namedref{Figure}{#1}}
\newcommand{\appendixref}[1]{\namedref{Appendix}{#1}}

\newcommand{\retriever}[1]{\textsc{c-REALM}}

\newcommand\blfootnote[1]{%
  \begingroup
  \renewcommand\thefootnote{}\footnote{#1}%
  \addtocounter{footnote}{-1}%
  \endgroup
}

\begin{document}
\maketitle
\begin{abstract}

The task of \emph{long-form question answering} (LFQA) involves retrieving documents relevant to a given question and using them to generate a paragraph-length answer. While many models have recently been proposed for LFQA, we show in this paper that the task formulation raises fundamental challenges regarding evaluation and dataset creation that currently preclude meaningful modeling progress. To demonstrate these challenges, we first design a new system that relies on sparse attention and
contrastive retriever learning to achieve state-of-the-art performance on the \emph{ELI5} LFQA dataset. While our system tops the public leaderboard, a detailed analysis reveals several troubling trends: (1) our system's generated answers are not actually grounded in the documents that it retrieves; (2) ELI5 contains significant train / validation overlap, as at least 81\% of ELI5 validation questions occur in paraphrased form in the training set; (3) ROUGE-L is not an informative metric of generated answer quality and can be easily gamed; and (4) human evaluations used for other text generation tasks are unreliable for LFQA. We offer suggestions to mitigate each of these issues, which we hope will lead to more rigorous LFQA research and meaningful progress in the future.\blfootnote{* Work done during an internship at Google Research.}\footnote{Resources accompanying our paper can be found in \url{https://github.com/martiansideofthemoon/hurdles-longform-qa}}

\end{abstract}
\section{Introduction}

Long-form question answering (LFQA) integrates the \emph{retrieval} component of open-domain QA, which involves searching a large external knowledge source for documents relevant to a given question, with a text \emph{generation} component to produce paragraph-length answers. Significant progress has been made on open-domain QA datasets such as Natural Questions~\citep{kwiatkowski2019natural}, whose questions are answerable with short phrases and entities, by leveraging dense retrieval techniques like ORQA~\citep{lee-etal-2019-latent}, REALM~\citep{guu2020realm}, and DPR~\citep{karpukhin2020dense,lewis2020retrieval,izacard2020leveraging}. Methods inspired by these results have recently been combined with pretrained language models~\citep{lewis2019bart,petroni2020kilt} and applied to the Reddit-derived ``Explain Like I'm Five'' (ELI5) dataset~\citep{fan-etal-2019-eli5}, which is the only publicly-available large-scale LFQA dataset.

The recently proposed KILT benchmark~\citep{petroni2020kilt}, which compares retrieval-augmented models across a variety of knowledge-intensive tasks including ELI5, automatically evaluates LFQA models by the quality of both generated answers (ROUGE-L against reference answers) and retrieved documents (R-precision against human-annotated relevant documents). In this paper, we build a state-of-the-art system\footnote{State-of-the-art as of April 3, 2021 --- the ``Google Research \& UMass Amherst'' team entry on \small{\url{https://evalai.cloudcv.org/web/challenges/challenge-page/689/leaderboard/1908}}} for ELI5 by using a sparse Transformer variant~\citep{roy2020efficient} to condition over Wikipedia paragraphs returned by a REALM-style retriever~\citep{guu2020realm}.

\begin{figure*}[t]
\includegraphics[width=1\linewidth]{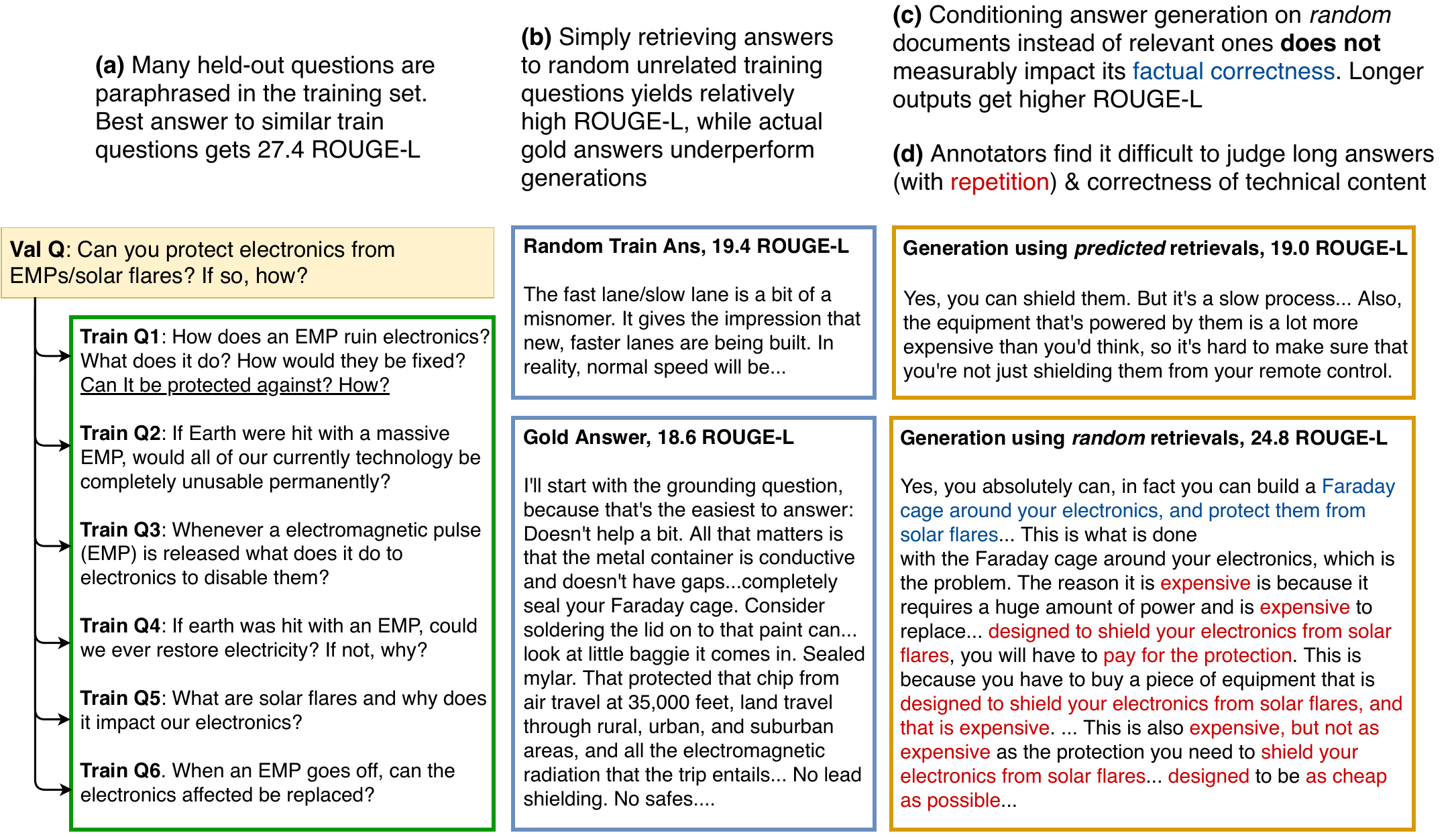}
\caption{A summary of the major hurdles (a-d) to progress in long-form question answering with ELI5.}
\label{fig:main_diagram}
\end{figure*}

However, despite its success on the KILT leaderboard, our system does not actually \emph{use} the documents that it retrieves! To measure the effect of retrieval on generation quality, we design a control experiment in which retrieved documents are replaced with randomly-sampled documents at inference time. Results from both human A/B tests and automatic metrics like ROUGE-L demonstrate that conditioning on random documents has almost no effect on generated answer quality (\figureref{fig:main_diagram}c). We recommend that future LFQA research report the results of such control experiments in addition to reporting generation and retrieval quality.

How can a system using random retrieval perform well on ELI5? Our analysis reveals that this result is partially due to significant train / validation overlap in the ELI5 dataset (\figureref{fig:main_diagram}a), which eliminates the need for external retrieval. A human study shows that at least 81\% of validation questions have a paraphrase in the training set, and almost all validation questions are topically similar to a training set question. While~\citet{fan-etal-2019-eli5} attempted to identify and remove question overlap using TF-IDF similarity, more complex semantic matching methods \& human verification is needed to address this issue in future LFQA datasets.

Digging deeper, we identify fundamental issues with using ROUGE-L to evaluate generated answer quality (\figureref{fig:main_diagram}b). Simple baselines such as just repeatedly copying the question, or choosing a random training set answer, can outperform  LFQA systems such as RAG~\citep{lewis2020retrieval} in terms of ROUGE-L. On the other hand, our system achieves \emph{higher} ROUGE-L than reference human-written answers, which is misleading since human A/B testers strongly prefer reference answers to our system's. We conclude that ROUGE-L is not a reliable metric to evaluate LFQA due to its large and relatively unconstrained output space (e.g., compared to translation or summarization), and we offer suggestions for better automatic \& human evaluations to enable meaningful progress on this task.

\section{A state-of-the-art LFQA system}\label{sec:model}
The ELI5 task~\citep{fan-etal-2019-eli5} asks models to generate paragraph-length answers to open-ended questions in English that often rely on world knowledge (e.g., \textit{how do jellyfish function without brains or nervous systems?}). LFQA systems thus benefit from conditioning answer generation on relevant documents from the web (such as the Wikipedia article about \textit{jellyfish}). While large-scale pretrained language models store surprising amounts of world knowledge within their parameters~\citep{petroni2019language, roberts2020much}, external document retrieval not only augments this intrinsic knowledge but also grounds model outputs in a knowledge source, which provides interpretability.

In this section, we describe our proposed LFQA system, which conditions answer generation on Wikipedia articles identified by a pretrained retriever. We use a dense retriever trained by scaling up a distantly supervised algorithm from~\citet{jernite2020}. Since retrieved articles can be quite long and often exceed the maximum sequence length of pretrained models like BERT~\citep{devlin-etal-2019-bert}, we use a sparse-attention variant of the Transformer to allow modeling over longer sequences. While our system sets a new state-of-the-art on ELI5, we question the significance of this result in \sectionref{sec:analysis_main}.

\subsection{Retriever}
We begin by specifying our dense retriever (``contrastive REALM'' or~\retriever~), which returns documents related to an input question.
Consider a corpus of long-form questions and answers, represented by $(q_i, a_i)^N_{i=1}$. Our retriever uses $q_i$ as a query to retrieve $K$ documents $(r_{i,j})_{j=1}^K$ from a knowledge corpus (Wikipedia), which is enabled by an encoder network that projects both questions and candidate documents to a 128-$d$ shared embedding space. Like REALM~\citep{guu2020realm}, our encoder is a BERT-base Transformer~\citep{devlin-etal-2019-bert} with a final projection layer. 

Since the ELI5 dataset does not include gold retrievals, we train our retriever by scaling up a method recently introduced by~\citet{jernite2020} that uses gold answers for distant supervision. The key idea is to push the encoded vector for a question close to a vector representation of its ground-truth answer(s), but away from all other answer vectors in the mini-batch (negative examples). Intuitively, this method works because both ELI5 answers and external documents are of paragraph length (documents are paragraph-length chunks from Wikipedia). Concretely, we optimize the loss,

\begin{align*}
    \text{loss} &= - \sum_{(q_i, a_i) \in B} \log \frac{\exp \mathbf{q}_i \cdot \mathbf{a}_i}{\sum_{a_j \in B} \exp \mathbf{q}_i \cdot \mathbf{a}_j}
\end{align*}

where $B$ is the mini-batch and $\mathbf{q}_i$, $\mathbf{a}_i$ are the encoded vector representations for $(q_i, a_i)$. This objective is based on contrastive learning, a method that has been used effectively for semi-supervised learning~\citep{chen2020simple} and dense retriever training~\citep{karpukhin2020dense}. Scaling up from~\citet{jernite2020}, who used a mini-batch size of 512 and initialized their retriever with BERT, we use much large mini-batches of size 12,288 (and hence, many more negative examples) and initialize our retriever with a strong pretrained retriever, the REALM model~\citep{guu2020realm} trained on the Common Crawl News (CC-News) corpus. These design decisions greatly improve retriever quality, as we observe in an ablation study (see \appendixref{appendix:ablation_batch_size}). During inference, we perform a maximum inner-product search (MIPS) with the ScaNN library~\citep{avq_2020} to efficiently find the top $K$ documents. In all our experiments we use $K=7$, following the setup in~\citet{guu2020realm}.

\subsection{Generator}

We next describe our generator model, which conditions its generated answers on retrieved documents returned by \retriever~. We use the Routing Transformer (RT) from~\citet{roy2020efficient}, which is the current
state-of-the-art in long-form language modeling. The RT is a sparse attention model that employs 
local attention as well as mini-batch $k$-means clustering to better model long-range 
dependencies in sequences (attention maps in \appendixref{appendix:training_model}). Long-form language models such as RT are well-suited to ELI5 as the task requires conditioning answer generation not only on a short question but also many lengthy retrieved documents. 

We pretrain our RT model on PG-19, a long-form language
modeling benchmark~\citep{rae2020compressive} created from approximately 28,000 Project Gutenberg books published before 1919. PG-19 has 1.9B tokens and an average context size of 69K words. While this data is out-of-domain for ELI5, we choose it to encourage long \& coherent generation. Our RT is a 22-layer model with 1032 hidden units (486M parameters), maximum sequence length of \(8192\) tokens, and a vocabulary of 98K subwords.\footnote{Our hyperparameters have been chosen manually with minimal tuning. See \appendixref{appendix:training_model} for details.}
We fine-tune our model in a decoder-only 
fashion~\citep{liu2018generating, wolf2019transfertransfo} by concatenating the top $K$ retrieved documents to the question
$[r_{i,K},~r_{i,K-1}~...~r_{i,1},~q_i,~a_i]$ and training the model to predict tokens of the answer $a_i$. We do not backpropagate gradients through the 
retriever.\footnote{We tried training the retriever jointly with RT using the attention bias scheme proposed in MARGE~\citep{lewis2020pre}. This 
improved perplexity only in \textit{autoencoding} settings where the gold answer itself is used as 
a retrieval query (like the setup in~\citealp{lewis2020pre}), which is not valid in LFQA.} Retrievals slightly improve perplexity (18.1 vs 17.8) as seen in~\citet{wang2020fly}, but do not improve generations (\S \ref{sec:grounding}).

\subsection{Main Experiments}
\label{sec:experiments}

\noindent \textbf{Dataset \& Evaluation details}: We evaluate our model on the KILT validation \& test subsets of ELI5~\citep{petroni2020kilt}, since the original ELI5 dataset does not have human annotations to measure retriever performance. We downloaded the ELI5 dataset~\citep{fan-etal-2019-eli5} from the KILT Github repository.\footnote{\url{github.com/facebookresearch/KILT}} This version of the dataset has 272,634 training examples, 1,507 validation examples and 600 test examples. The test set answers are hidden, and hosted on a public leaderboard in the EvalAI platform~\citep{EvalAI}.

Answer quality is measured by the maximum overlap of generations with a set of gold answers in terms of unigram F1 score and ROUGE-L~\citep{lin-2004-rouge}.~\citet{petroni2020kilt} collected human annotations of Wikipedia articles which support 
ELI5 gold answers, which enables measuring retrieval quality by computing R-precision (if
the top-1 retrieval matches the annotation) and Recall@5 using the top-5 
retrievals. Finally, the KILT benchmark combines R-prec. and 
ROUGE-L to measure the overall performance of the system by ``KILT ROUGE-L''. This metric is similar to ROUGE-L, but assigns a score of 0 whenever the top-1 retrieval does not match the gold annotation.\\

\noindent \textbf{Baselines}: We compare our model with the other entries on the ELI5 KILT leaderboard which are either generation-only, like T5-base~\citep{raffel2019exploring} and BART~\citep{lewis2019bart}, or variants of BART using retrieval such as RAG~\citep{lewis2020retrieval} and BART + DPR~\citep{petroni2020kilt}. These systems are based on massive pretrained language models, with similar number of parameters as our model (details in \appendixref{appendix:number_parameters}).\\

\begin{table}[t]
\small
\begin{center}
\begin{tabular}{ lrrrrr } 
 \toprule
 & \multicolumn{2}{c}{\emph{Retrieval}} & \multicolumn{2}{c}{\emph{Generation}}\\
Model & RPr. & R@5 & F1 & R-L & KRL \\
\midrule
T5-base & 0.0 & 0.0 & 16.1 & 19.1 & 0.0 \\
BART & 0.0 & 0.0 & 19.2 & 20.6 & 0.0 \\
RAG & \textbf{11.0} & 22.9 & 14.5 & 14.1 & 1.7 \\
BART + DPR & 10.7 & \textbf{26.9} & 17.9 & 17.4 & 1.9 \\
\midrule
$p = 0.9$ \\
RT + REALM & 6.7 & 15.5 & 25.1 & 21.5 & 1.4 \\ 
RT + \retriever & 10.2 & 24.4 & \textbf{25.4} & 21.5 & 2.1\\\\

$p = 0.6$ \\
RT + REALM & 6.7 & 15.7 & 23.1 & \textbf{23.4} & 1.5\\ 
RT + \retriever & 10.7 & 24.6 & 22.9 & 23.2 & \textbf{2.4} \\
\bottomrule
\end{tabular}
\end{center}
\caption{Results on the KILT test set for ELI5 for (1) retrieval performance, using R-precision and Recall@5 (RPrec, R@5), and (2) generation quality, using ROUGE-L (R-L). These scores are combined to produce the final metric KILT R-L (KRL). We outperform prior work on both generation \& combined scores.}
\label{tab:kilt_test_main_results}
\end{table}

\noindent \textbf{Results}:  \tableref{tab:kilt_test_main_results} contains our results on the test set of the ELI5 (also on the public KILT leaderboard). We present four variants of our system, using a different retriever during inference (REALM or~\retriever~), and different nucleus sampling $p$ values~\citep{holtzman2020curious}. All variants outperform prior work in generation quality,
with lower-entropy models ($p=0.6$) performing best.\footnote{As in~\citet{holtzman2020curious}, a human study reveals that higher entropy ($p=0.9$) answers are slightly more coherent and sensible, but lower entropy answers ($p=0.6$) are more relevant to the question (details in \appendixref{appendix:human_evaluation}).} ~\retriever~ performs competitively to RAG and DPR despite being only distantly supervised, and outperforms REALM. Our proposed RT+\retriever~ system achieves a new state-of-the-art on combined performance (KILT R-L). Generations from our model are provided in \figureref{fig:generations_main} and \appendixref{appendix:generations}.

\begin{figure}[t]
    \footnotesize
    \centering
    \begin{tabular}{p{0.95\columnwidth}}
    \toprule
  Q: Why are almost all boats white?\\
  \midrule
  A: Boats are generally painted white for obvious reasons, as they are inexpensive and easy to clean. Ships are also often painted to provide a visible visual contrast to the sea. Boats also often have high visibility ports to allow for easy international trade. In addition, many boats have an air-breathing function that makes them easier to spot and stay afloat, which is a good thing if you're a pirate or trying to get into some country's waters. Lastly, many ships are built in the last few decades, so they are not as exotic looking as they were thirty years ago. And lastly, many ships are made out of materials that are easy to work with, and so many of them are painted white as a practical matter. \hl{There are many, many more reasons for why ships are white, but those are the ones I know of. I'm sure there are many other reasons, but those are the ones I know of.} Edit: I forgot to mention that the other answers about visibility and easy to see ports are also true. Ships are painted to appear to be larger, and more expensive. That makes them appear more impressive, and so they are more likely to be bought by people who want to show off their wealth and status.\\
  \bottomrule
       \end{tabular}
    \caption{Example generation from our LFQA system with \(p=0.9\). Generations are long \& coherent, but suffer from \hl{repetition} towards the end. (more in \appendixref{appendix:generations} and attached data supplementary material).}
    \label{fig:generations_main}
\end{figure}

\section{Analysis}
\label{sec:analysis_main}

In this section, we conduct a thorough analysis of our model's usage of retrievals (\sectionref{sec:grounding}), the impact of overlap in ELI5's train / validation / test folds (\sectionref{sec:train_valid_overlap}), issues with ROUGE-L and performance bounds (\sectionref{sec:rougel_bounds}), and the difficulty in human evaluation for this task (\sectionref{sec:human_eval_difficulty}). At the end of each section, we provide short takeaways with suggestions for future work.
\subsection{Are generations grounded in retrieval?}
\label{sec:grounding}

While our retrieval-augmented system achieves state-of-the-art performance, we find little evidence that it is actually \emph{using} the retrieved documents. To measure this, we run an ablation study where at \emph{inference time} we replace retrieved paragraphs with randomly sampled paragraphs from Wikipedia. We compare this \emph{Random} baseline with our original system (\emph{Predicted}) in terms of generation quality as well as the $n$-gram overlap between the generation and the retrieved paragraphs.\\

\begin{table}[t]
\small
\begin{center}
\begin{tabular}{ lrrrrrr } 
 \toprule
   &  & \multicolumn{2}{r}{vs predicted retr.} & \multicolumn{2}{c}{vs random retr.} \\
 & R-L & ~~~~~~~~~1-g & 2-g & ~~~~~~~1-g & 2-g \\
\midrule
\emph{Predicted}  & 24.42 & 52.3 & 9.0 & 38.8 & 3.9 \\
\emph{Random} & 24.20 & 51.2 & 8.5 & 38.5 & 3.9 \\
\midrule
\emph{Gold Ans} & - & 54.1 & 9.1 & 40.2 & 3.8 \\
\bottomrule
\end{tabular}
\end{center}
\caption{Comparison of generations (with $p=0.6$) conditioned on predicted retrievals (\emph{Predicted}) and randomly chosen retrievals (\emph{Random}). Notice small differences in: (1) ROUGE-L vs gold answers (R-L); (2) $n$-gram overlap ($n$-g) with predicted retrievals (vs predicted retr.). Gold answers also have a similar overlap with predicted retrievals. To control for stopwords, we show overlaps with the random retrievals.}
\label{tab:pred_vs_random_all}
\end{table}

\begin{table}[t]
\small
\begin{center}
\begin{tabular}{ lllrrr } 
 \toprule
A & B &  Prefer A & Prefer B & Tie \\
\midrule
\multicolumn{5}{l}{For $p=0.6$}\\
 pred. & random & 40\% (78) & 33\% (~~64) & 27\% (51) \\
 pred. & gold ans. & 14\% (29) & \textbf{68}\% (138) & 18\% (36) \\
 \midrule
\multicolumn{5}{l}{For $p=0.9$}\\
  pred. & random & 31\% (52) & 37\% (~~63) & 32\% (54)\\
 pred. & gold ans. & 17\% (49) & \textbf{72}\% (203) & 11\% (31) \\
\bottomrule
\end{tabular}
\end{center}
\caption{Human evaluation results with exact number of ratings shown in ($\cdot$). Annotators are shown a question along with two answers (A, B) in random order and ask them to choose one (details in \appendixref{appendix:human_evaluation}). For both model variants ($p=0.6, 0.9$), we see (1) little difference between generations conditioned on predicted (pred.) or random (rand.) retrievals; (2) strong preference for gold answers over generations. }

\label{tab:human_evaluation}
\end{table}

\begin{table}[t]
\small
\begin{center}
\begin{tabular}{ lrrr } 
 \toprule
  & vs qn. & vs predicted retr. & vs random retr. \\
 & & but not in qn. & but not in qn. \\
\midrule
\multicolumn{4}{l}{(lemmatized nouns, proper nouns, numbers only)} \\\\
\emph{Predicted} & 13.4\% & 34.4\% & 11.9\% \\
\emph{Random} & 13.7\% & 31.7\% & 12.1\% \\
\midrule
\emph{Gold Ans} & 8.3\% & 28.8\% & 15.1\% \\
\bottomrule
\end{tabular}
\end{center}
\caption{A fine-grained version of \tableref{tab:pred_vs_random_all} measuring the unigram overlap of nouns/numbers in the generations with the input question (vs qn.), retrievals predicted by \retriever~ (vs predicted retr.) and randomly sampled retrievals (vs random retr.). Similar to \tableref{tab:pred_vs_random_all}, notice very little difference with and without retrieval.}
\label{tab:pred_vs_random_all_nouns}
\end{table}

\noindent \textbf{Generations are similar irrespective of type of retrievals}: We present our results in \tableref{tab:pred_vs_random_all}. Despite not being conditioned on any meaningful retrievals, the \emph{Random} retrieval model has similar ROUGE-L scores as our \emph{Predicted} system. Moreover, generations from the \emph{Random} and \emph{Predicted} models have similar amounts of 1-gram and 2-gram overlap with the paragraphs retrieved by \retriever~, despite the fact that the \emph{Random} model does not actually see the retrieved paragraphs.\footnote{Corresponding experiments with the $p=0.9$ variant of our model are presented in \appendixref{appendix:grounding_correct_retrievals}.}

The $n$-gram overlaps are possibly overestimates due to stopwords (e.g., prepositions, punctuation) and entities which are copied from the question. To tackle this issue, in \tableref{tab:pred_vs_random_all_nouns} we measure the fractions of lemmatized nouns, proper nouns and numbers in the generated answer which are present in the predicted retrievals but not in the question. We notice similar trends as before, with only small differences between the two systems.
Finally, there is almost no correlation (Spearman $\rho = 0.09$) between the \emph{Predicted} model's generation quality and the amount of unigram overlap between its outputs and the retrieved documents (scatter plots in \appendixref{appendix:grounding_correct_retrievals}), strengthening our hypothesis that generations are not grounded in retrievals.\footnote{All these trends persist even on questions for which our retriever predicts the ground-truth document (\appendixref{appendix:grounding_correct_retrievals})}\\

\noindent \textbf{Human evaluation validates our findings}: As ROUGE-L and $n$-gram overlap have major limitations for LFQA (\sectionref{sec:rougel_bounds}), we perform additional human A/B testing on the output of \emph{Random} and \emph{Predicted}. Specifically, we ask human volunteers\footnote{Details of our experimental setup in \appendixref{appendix:human_evaluation}.} to choose between answers generated by the two systems (presented in random order). As seen in \tableref{tab:human_evaluation}, humans struggle to choose which of the two answers is more relevant to the question. For both model variants ($p=0.6, 0.9$), there is a less than 7\% preference for a particular answer type, with humans preferring answers (by 6\%) from the \emph{Random} model for $p=0.9$!\\

\noindent \textbf{Other systems also have this issue, possibly due to source-reference divergence and train-validation overlap}: We note that this issue is not unique to our system --- other systems on the KILT leaderboard like BART + DPR and RAG actually perform \emph{worse} than their no-retrieval counterpart (BART) in generation quality, as shown in \tableref{tab:kilt_test_main_results}. Qualitatively, we found no evidence of retrieval usage in a publicly hosted ELI5 model demo by~\citet{jernite2020}.\footnote{\url{https://huggingface.co/qa}} A possible explanation for this issue is high source-reference divergence, a common problem in table-to-text generation~\citep{wiseman2017challenges, tian2019sticking}. In \tableref{tab:pred_vs_random_all} and \tableref{tab:pred_vs_random_all_nouns}, we measure the $n$-gram overlap of top-ranked gold validation answers (\emph{Gold Ans}) with predicted retrievals. This overlap is low and similar to that of our generations, which we suspect encourages our model to ignore retrievals. A second explanation is the large amount of train-validation overlap (\sectionref{sec:train_valid_overlap}), which eliminates the need for retrieval.\\

\noindent \textbf{Why does our model do well compared to other systems despite not using retrievals?} While our model has similar capacity as the BART/RAG baselines (comparison in \appendixref{appendix:number_parameters}), we hypothesize that our improvements in ROUGE-L are due to a different pretraining objective. BART is pretrained on a masked infilling task on short sequences. Instead, we pretrain our model to perform next-word prediction on long sequences from Project Gutenberg,  which encourages long \& fluent generations. To illustrate this length effect, in \appendixref{appendix:length_effect} we show that truncated outputs from our model get lower ROUGE-L scores on ELI5.\footnote{While we do not have access to generations from baselines on the KILT leaderboard, example generations from the demo of the BART model in ~\citet{jernite2020} are significantly shorter (59 words avg.) than our generations (187 words avg.).} Prior summarization literature~\citep{sun2019compare} has also shown that ROUGE scores vary heavily by length. To compare the same systems on shorter length outputs, we also tried finetuning the pretrained model on Wizard of Wikipedia~\citep{dinan2019wizard}, an unconstrained dialogue generation task with \textbf{single sentence} dialogues (much shorter than ELI5). As seen on the public KILT leaderboard,\footnote{\url{https://eval.ai/web/challenges/challenge-page/689/leaderboard/1909}} our system has \emph{lower} ROUGE-L scores than the BART / RAG baselines. Another possible explanation is issues with ROUGE-L itself, as discussed in \sectionref{sec:rougel_bounds}.\\

\noindent \textbf{Takeaway (better evaluation of grounding)}: For evaluating LFQA, it is important to run control experiments with random retrievals \& measure grounding of generations in retrieval. While the KILT benchmark does attempt to measure the combined retrieval + generation performance via KILT RL, it does not check whether the generations actually \emph{used} the retrievals. In other words, one can submit independent retrieval \& generation systems, but still perform well on the combined score. This may not be an issue for short-form QA tasks like Natural Questions, since the gold answer is often exactly contained as a span in the gold retrieval. Also, as retrieval might be less important for large language models with parametric knowledge~\citep{roberts2020much}, the KILT-RL strategy of simply aggregating top-1 retrieval score with ROUGE-L unfairly penalizes systems not relying on retrieval.\footnote{Another issue of KILT-RL is ignoring non top-1 retrievals, penalizing models using multiple retrievals together in context.}

\subsection{Training / Validation Overlap}
\label{sec:train_valid_overlap}

Our experiments in \sectionref{sec:grounding} show that model performance is mostly unchanged by 
conditioning generation on randomly sampled retrievals instead of predictions from~\retriever~. 
Despite not using retrievals, we observe qualitatively that our model displays a large amount of parametric knowledge (``Faraday Cage'' in \figureref{fig:main_diagram}c), which is surprising since 
it was pretrained on novels from Project Gutenberg (not Wikipedia). In this section, 
we discover that a major reason for ignoring retrievals is the large amount of train / validation overlap in ELI5. While ~\citet{fan-etal-2019-eli5} attempted to fix this issue through TF-IDF overlap, this method is insufficient to identify all question paraphrases, as we find significant overlap between the training set and the KILT validation set of ELI5.\footnote{The ELI5 demo from~\citet{jernite2020} also retrieves the top-1 similar training set question. Qualitatively, we found many validation examples had near-identical train paraphrases.} ELI5 is not the only dataset with substantial train / test overlap: ~\citet{lewis2020question} identify similar issues with short-form QA datasets like Natural 
Questions.\\

\noindent \textbf{Finding similar questions \& measuring overlap}: We use our retriever~\retriever~~to \emph{retrieve} similar questions from the training set, since it has learned to map questions to a feature-rich embedding space. For each validation question, we retrieve the 7 most similar training set questions. We use both human and automatic evaluation to calculate the amount of overlap. For human evaluation, we show annotators on Amazon Mechanical Turk\footnote{We pay workers 4 cents per question pair (\$8-12 / hr). We only hire workers
from USA, UK and Australia with a 95\% or higher approval rating and at least 1000 approved HITs.} a validation set question and a retrieved training set question, and ask them to annotate the pair as \textbf{0}: No paraphrase relationship; \textbf{1}: on similar topics, but different questions; \textbf{2}: approximately the same question (an adaptation of the paraphrase evaluation of~\citealp{kok2010hitting}). We take 300 validation set questions and ask three crowd-workers to rate them against retrieved training questions on this scale, and consider the label with majority rating. To improve quality, we manually verify their annotations. 

\tableref{tab:human_qa_overlap} shows that \textbf{81\%} of validation set questions have at least one paraphrase in the training set, while \textbf{all} annotated questions have at least one topically similar question in the training set, which indicates substantial training / validation overlap. The experiment had ``fair agreement'' with a Fleiss $\kappa$ of 0.29~\citep{fleiss1971measuring, landis1977measurement}.

\begin{table}[t]
\small
\begin{center}
\begin{tabular}{ lrr } 
 \toprule
qns with at least one train set paraphrase & 81\% \\
qns with at least one train set topically similar & 100\% \\
\midrule
\% of all pairs marked paraphrases & 39.5\% \\
\% of all pairs marked topically similar & 47.8\% \\
\% of all pairs marked as non-paraphrases & 12.7\% \\
\bottomrule
\end{tabular}
\end{center}
\vspace{-0.1in}
\caption{A human evaluation measuring the amount of overlap between validation set questions (qns) and retrieved questions from the training set.}
\vspace{-0.1in}
\label{tab:human_qa_overlap}
\end{table}

As manually annotating question overlap can be expensive and time-consuming, we also experiment with automatic overlap detection methods. In particular, we use a RoBERTa-large binary classifier~\citep{liu2019roberta} fine-tuned on the Quora Question Paraphrase (QQP) dataset~\citep{iyer2017quora} from the GLUE benchmark~\citep{wang2019glue}. For 43.6\% of the ELI5 validation set, this classifier marked at least one retrieved question as a paraphrase (46\% for the 300 questions we annotated). Qualitatively, we notice that this classifier often mis-classifies retrieved questions that are valid paraphrases but exhibit significant lexical or syntactic divergence. This observation, along with the smaller fraction of valid paraphrases in the QQP training set (37\%), partially explains the gap between automatic \& human evaluations. \\

\noindent \textbf{Using retrieved QA for generation}: Since ELI5 contains significant amount of overlap between the training and validation sets, a system can simply copy the answers of retrieved training set questions instead of actually doing generation.~\tableref{tab:rouge_bounds} shows that by using the longest answer within the top-$K$ retrieved questions,
we outperform two prior systems (RAG, BART + DPR) that use retrieval-augmented generation. As an upper bound, we also consider a system which uses the best possible answer to retrieved training set questions in terms of ROUGE-L (\emph{best top-K train answer}). This system gets 28.5 ROUGE-L, outperforming all others. \\

\noindent \textbf{ELI5 performance on overlapping QA}: Finally, we measure the performance difference between validation questions that overlap with the training set vs. those that do not. Since we only have human annotations for 300 questions (the no-overlap subset has only 53 samples), we present this analysis using the QQP classifier's outputs as well. In \tableref{tab:overlap_vs_no_overlap}, we notice large differences of 6.6 RPrec, 8.1 R@5 in retrieval performance favoring the overlap subset, but only a small generation score gain of 0.8 F1, 0.4 R-L (which may be misleading as discussed in \sectionref{sec:rougel_bounds}).\\

\begin{table}[t]
\small
\begin{center}
\begin{tabular}{ lrrrr } 
 \toprule
 & \multicolumn{2}{c}{Retrieval} & \multicolumn{2}{c}{Generation} \\
Split & ~~~RPrec & R@5 & F1 & R-L \\
\midrule
\multicolumn{5}{l}{QQP classifier (1.5k examples)}\\\\
overlap (43.6\%) & \textbf{17.0} & \textbf{25.8} & \textbf{26.0} & \textbf{24.6} \\
not overlap (56.4\%) & 10.4 & 17.7 & 25.2 & 24.2 \\
\midrule
\multicolumn{5}{l}{AMT evaluation (300 examples)}\\\\
overlap (81\%) & \textbf{14.0} & \textbf{20.0} & \textbf{25.0} & 24.3 \\
not overlap (19\%) & 5.3 & 17.9 & 24.5 & \textbf{24.8} \\
\bottomrule
\end{tabular}
\end{center}
\vspace{-0.1in}
\caption{ELI5 performance difference (for the $p=0.6$ model) between subsets of validation QA having a question paraphrase (overlap) and not having a question paraphrase (not overlap) in the training set. We see the overlap 
subset has much better retrieval performance and slightly better generation performance.}
\vspace{-0.1in}
\label{tab:overlap_vs_no_overlap}
\end{table}

\noindent \textbf{Takeaway (careful held-out curation)}: Based on our findings, we suggest that more careful dataset curation for LFQA tasks is needed to prevent duplicates. While we acknowledge the efforts of~\citet{fan-etal-2019-eli5} to fix this issue, we also suggest alternative methods to control overlap and focus on evaluating generalization in held-out sets: (1) automatically retrieving paraphrases and then running human validation to eliminate them; or (2) holding out entire genres or domains to reduce the possibility of overlap --- for example, keeping Q/A on \textit{Sports} only in the held-out sets. Note that simply pruning the existing splits using these criteria will significantly reduce the size of the held-out datasets; so we suggest re-splitting the train/validation/test splits from the entire pool of collected questions.

\subsection{ROUGE-L Bounds on ELI5 Performance}
\label{sec:rougel_bounds}

We have seen that simply copying the answer of a close question paraphrase from the training set achieves 28.5 ROUGE-L with an optimal selection among retrieved questions and outperforming all computational models. But how ``good'' is this absolute number? What are some suitable upper \& lower bounds to ROUGE-L scores on ELI5? Is ROUGE-L an informative metric for LFQA?\\

\noindent \textbf{Lower bounds} are trivial baselines used to test the vulnerability of datasets or metrics to simple heuristic strategies that do not actually perform the task. Recent examples include hypothesis-only baselines for natural language inference~\citep{gururangan2018annotation} and passage-only baselines for reading comprehension~\citep{kaushik2018much}. We evaluate two ROUGE-L lower bounds on ELI5:

\noindent (1) copy the question 5 times and concatenate, as longer outputs boost ROUGE-L (\appendixref{appendix:length_effect}); \\
\noindent (2) retrieve a \textit{random} training set answer.

Our first baseline contains entities often present in the gold answer, but without actually answering the question. Our second baseline follows the ``style'' of an answer but is completely off-topic. \\

As an \textbf{upper bound}, we estimate the ROUGE-L of gold answers themselves. On an average, there are 12 gold answers per question, so we measure the ROUGE-L of the longest gold answer with respect to the other gold answers. We also measure the maximum pairwise ROUGE-L between two gold answers for the same question.\footnote{Note that different gold answers were not written independently as Reddit users writing answers can read existing answers and may want to provide a non-overlapping perspective. Due to the high train/valid overlap, the \emph{best top-7 retrieved answer} could be a better upper bound since it is from another Reddit post (and performs better than \emph{best gold answer}).
} We only calculate upper bounds for the validation set, since the gold answers of the KILT test set are hidden.\\

\begin{table}[t]
\small
\begin{center}
\begin{tabular}{ lrrrr } 
 \toprule
 & \multicolumn{2}{c}{Validation}  & \multicolumn{2}{c}{Test} \\
 Scheme & F1 & R-L & F1 & R-L \\
 \midrule
 random train answer $(\downarrow)$ & 17.8 & 16.2 & 17.1 & 15.5 \\
 copy input $(\downarrow)$ & 16.6 & 20.0 & 14.8 & 16.9 \\
 \midrule
 RAG~\shortcite{lewis2020retrieval} & 17.2 & 16.1 & 14.5 & 14.1 \\
 BART + DPR~\shortcite{petroni2020kilt} & 18.8 & 18.5 & 17.9 & 17.4 \\
longest top-1 train answer & 25.2 & 20.7 & 21.6 & 18.7 \\
longest top-7 train answer & 26.9 & 21.1 & 22.0 & 18.5 \\
 RT +~\retriever~ (ours) & 25.6 & 24.4 & 22.9 & 23.2\\ 
 \midrule
 best top-1 train answer $(\uparrow)$ & 25.9 & 22.4 & - & - \\
best top-7 train answer $(\uparrow)$ &  31.5 & 28.5 & - & - \\
longest gold answer $(\uparrow)$ & 26.7 & 21.2 & - & - \\
best gold answer $(\uparrow)$ & 29.5 & 26.2 & - & - \\
\bottomrule
\end{tabular}
\end{center}
\caption{Upper $(\uparrow)$ and lower $(\downarrow)$ bounds to performance on ELI5. Lower bounds have been submitted to the public KILT leaderboard, as ``Metrics Test''.}
\label{tab:rouge_bounds}
\end{table}

\noindent \textbf{Lower bounds beat prior work, upper bounds have low ROUGE-L}: We compare our bounds with actual retrieval augmented generation systems in \tableref{tab:rouge_bounds}. Both our lower bounds (\emph{random training answer}, \emph{copy input}) are quite competitive, outperforming RAG~\citep{lewis2020retrieval} and performing close to BART + DPR~\citep{petroni2020kilt} without actually answering the question! This shows that ROUGE-L is fairly sensitive to simply copying entities from the question as well as stylistic properties of ELI5. On the other hand, upper bounds  (\emph{longest gold answer}) perform worse than our system (21.2 vs 24.4). Suspecting that this result is misleading, we run another human A/B test by showing volunteers a question and asking them to choose between answers generated by our system and the \emph{longest gold answer}, shuffled at random.\footnote{Human A/B testing details in \appendixref{appendix:human_evaluation}.} As seen in \tableref{tab:human_evaluation}, the majority of humans prefer the gold reference answers vs generations (68\% vs 14\% for $p=0.6$). In interviews with human annotators after completing the task, they reported that both answers were often fluent and stylistically similar, but one eventually veered off-topic.\\

\noindent \textbf{Takeaway (better automatic metrics needed)}: Our experiments demonstrate that computing the ROUGE-L of generations against gold answers is not a meaningful way to evaluate LFQA systems, since it is not selective enough to differentiate between valid/invalid answers. There is a very small margin of improvement between trivial lower bounds and strong upper bounds, with the absolute scores of upper bounds being quite low. We suspect this is due to the long length of answers and fairly unconstrained and large output space. The ELI5 dataset has several open-ended questions with many plausible answers (like \emph{What causes traffic?}), often involving analogies. A possible fix is a sentence-level evaluation and then aggregating scores across generated sentences, but appropriate penalties are needed for lack of diversity~\citep{zhu2018texygen} and short lengths. Other possible fixes include learning task-specific metrics to measure semantic overlap~\citep{sellam2020bleurt} or metrics to check factual correctness~\citep{zhang2019optimizing} and faithfulness to input~\citep{wang2020asking, durmus2020feqa, zhou2020detecting}. Ultimately, all automatic metrics have their limitations, and human evaluation is necessary~\citep{celikyilmaz2020evaluation}.

\subsection{Difficulty of Human Evaluation}
\label{sec:human_eval_difficulty}

To better understand the inherent difficulty of evaluation in ELI5, we interviewed human annotators (of~\tableref{tab:human_evaluation}) and found two challenges:\\


\noindent \textbf{(1) Unfamiliarity with question topics}: While most annotators found the Q/A interesting, they were often unfamiliar with the technical topics discussed in the questions. This made it hard for them to assess answer correctness. The ELI5 dataset has questions in a wide variety of topics~(History, Politics, Biology etc.), while most annotators were Computer Science graduate students. While we did allow annotators to use Wikipedia, they mentioned domain-experts will be better judges of factual correctness of answers.\\

\noindent \textbf{(2) Length of Answers}: Annotators mentioned the paragraph-long length of answers made the task quite challenging. Annotators reported taking an average of 2 minutes per answer pair, many of which required careful thought \& concentration. This was especially difficult when only part of the answer was correct and the rest had contradictions or repetitions, a common theme in our generations.\\

\noindent \textbf{Takeaway}: Human evaluation is challenging but necessary for evaluating LFQA. Crowd-workers are unlikely to spend time reading \& analyzing long text~\citep{akoury2020storium}. Hence, it is imperative to design \emph{simpler} evaluations. One effort in this direction is~\citet{dugan-etal-2020-roft}, who reveal one generated sentence at a time and estimate system quality based on the number of sentences which fooled humans. Another promising direction is extrinsic evaluation~\citep{celikyilmaz2020evaluation} where humans actually interact with systems in real-world scenarios such as the Alexa Prize~\citep{ram2018conversational} or STORIUM~\citep{akoury2020storium}.

\section{Conclusion}

We present a ``retrieval augmented'' generation system that achieves state-of-the-art performance on the ELI5 long-form question answering dataset. However, an in-depth analysis reveals several issues not only with our model, but also with the ELI5 dataset \& evaluation metrics. We hope that the community works towards solving these issues so that we can climb the right hills and make meaningful progress on this important task.

\section*{Acknowledgements}

First and foremost, we thank the twenty people who volunteered to help out with with the human annotation experiments. We are very grateful to Vidhisha Balachandran, Niki Parmar, and Ashish Vaswani for weekly meetings discussing progress and the REALM team (Kenton Lee, Kelvin Guu, Ming-Wei Chang and Zora Tung) for help with their codebase and several useful discussions which helped us improve our experiments. We are grateful to Tu Vu for help with the QQP classifier. We thank Jules Gagnon-Marchand and Sewon Min for suggesting useful experiments on checking ROUGE-L bounds. Finally, we thank Shufan Wang, Andrew Drozdov, Nader Akoury, Andrew McCallum, Rajarshi Das, and the rest of the UMass NLP group for helpful discussions and suggestions at various stages in the project. This work was primarily done during KK's internship at Google Brain, mentored by AR. MI and KK are supported by award IIS-1955567 from the National Science Foundation (NSF).

\section*{Ethical Considerations}

Our system faces a similar set of issues as most modern text generation technology, like fabrication of facts~\citep{zellers2019defending}, potential for misuse~\citep{brown2020language} and reflecting biases prevalent on Reddit (the ELI5 dataset has been built using the \texttt{r/ELI5} subreddit). In our work, we attempted to make text generators more factually grounded by conditioning generations on retrieved Wikipedia articles, hoping to reduce fact fabrication. Unfortunately, a thorough analysis (\sectionref{sec:grounding}) has revealed that our system is still not grounding its generations in retrievals, and we have recommended the design of better metrics to measure factual correctness to tackle this issue.

Our final models were trained using 64 Google Cloud TPUs for a total of 32 hours. As mentioned in the Google 2019 environment report,\footnote{\url{https://www.gstatic.com/gumdrop/sustainability/google-2019-environmental-report.pdf}} ``TPUs are highly efficient chips which have been specifically designed for machine learning applications''. These accelerators run on Google Cloud, which has ``matched 100\% of its electricity consumption with renewable energy purchases, and has committed to fully decarbonize its electricity supply by 2030'' (\url{https://cloud.google.com/sustainability}).
More details on training time are provided in \appendixref{appendix:training_model}.
\bibliography{bib/journal-full,bib/anthology,bib/custom}
\bibliographystyle{acl_natbib}
\newpage
~
\newpage
\appendix

\section{Appendices for ``Hurdles to Progress in Long-form Question Answering''}
\label{sec:appendix}

\subsection{Training \& Model Details}
\label{appendix:training_model}

All our models are developed and trained using TensorFlow 1.15~\citep{abadi2016tensorflow} and Tensor2Tensor~\citep{tensor2tensor}. Our implementations are based on the open-source codebases of REALM \footnote{\url{https://github.com/google-research/language/tree/master/language/realm}} and the Routing Transformer. \footnote{\url{https://github.com/google-research/google-research/tree/master/routing_transformer}} Similar to the REALM implementation, we use separate processes to run the retriever and generate training data (using a MIPS search). Since our retriever is frozen, we do not use the document index refresher available in their codebase.\\

\noindent \textbf{Retriever}: Our retriever is trained on 64 Google Cloud TPUs for a total of 4k steps and a batch size of 12288. We do early stopping on the validation data (with a smaller batch size of 512 due to smaller P100 GPU memory). Our model converges quite fast, reaching its best performance in 1.5k steps (in 43 minutes) and needing 103 minutes for the full set of 4k steps.\\

\noindent \textbf{Generator}: Our generator is trained on 64 Google Cloud TPUs, for a total of 100k steps on the ELI5 training set. We use the \texttt{pg19\_local\_cluster8k} configuration available in the Routing Transformer implementation. Besides the default hyperparameters, setting 15\% input, attention and ReLU dropout was critical to prevent overfitting on the training set. We use a learning rate of 5e-5. Our retrievals, questions and answers are truncated / padded to 288 subword tokens (using the PG19 subword tokenizer). We use a minibatch size of 128 QA pairs, which corresponds to 332k tokens per mini-batch (of which, the loss is computed over the last 288 answer tokens, or 37k total tokens). We do not compute loss over padded tokens, and use special symbols to separate different parts of the input context. We reverse the retrieved paragraphs in context since the model uses local attention layers, and we wanted higher ranked retrievals to appear closer to the answer tokens. Our models take about 30 hours to finish 100k steps (0.92 steps / second).\\

\begin{figure}[h]
\begin{subfigure}{.24\textwidth}
  \centering
  \includegraphics[width=.8\linewidth]{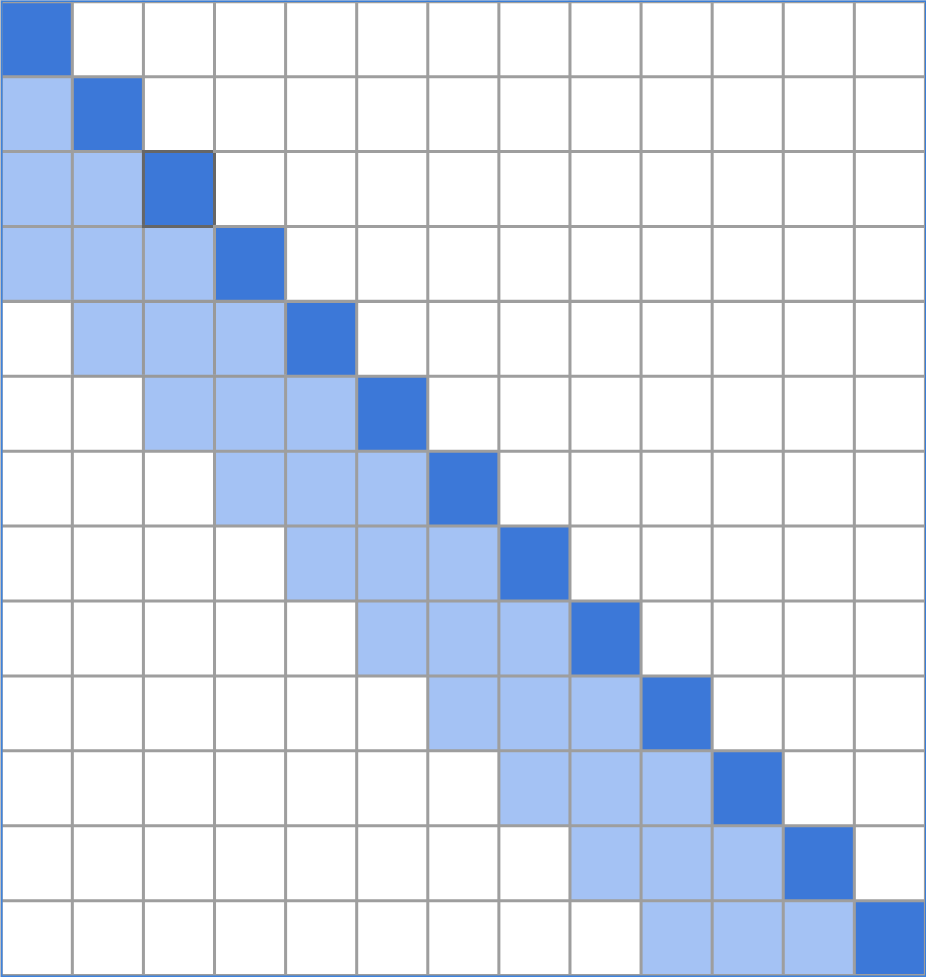}
  \caption{Local attention}
  \label{fig:sfig1}
\end{subfigure}%
\begin{subfigure}{.24\textwidth}
  \centering
  \includegraphics[width=.8\linewidth]{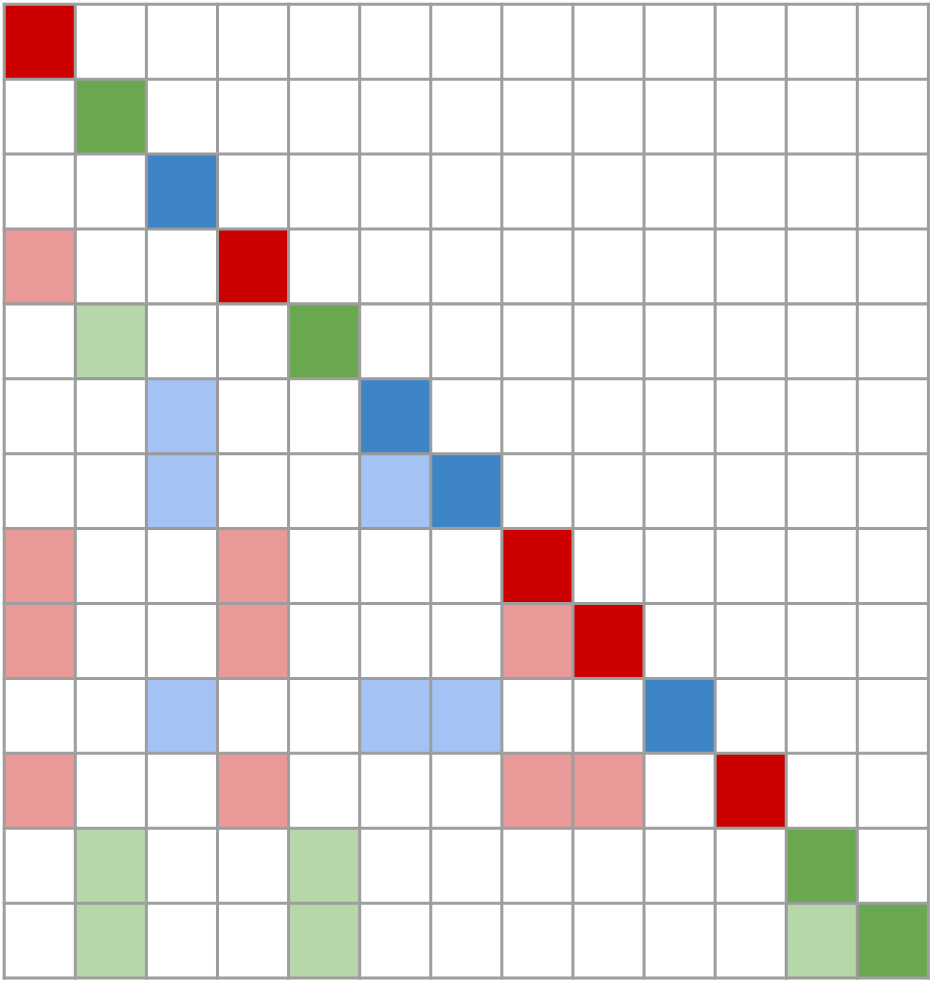}
  \caption{Routing attention}
  \label{fig:sfig3}
\end{subfigure}
\caption{Figures (from~\citealp{roy2020efficient}) showing 2-D attention schemes for the sparse attention mechanism used in
Routing Transformer. Lower layers pool in local information via sliding window local attention
(Sub-figure~\ref{fig:sfig1}) while upper layers gather global information for every token via clustering (Sub-figure~\ref{fig:sfig3}).}
\label{fig:attention}
\end{figure}

\noindent \textbf{Attention Maps}: We show the 2D plots of our generator's attention maps in \figureref{fig:attention}.

\noindent \textbf{Hyperparameter Choices}: We experimented with several different pretraining strategies (using Wikipedia), smaller model variants and hyperparameter choices \emph{manually} in preliminary experiments. All these experiments performed quite poorly on ELI5, producing very short and sometimes incoherent responses. Finally, switching to a Routing Transformer model which was pretrained on a longform language modeling dataset (PG-19) significantly improved generation quality. Hyperparameters for this pretrained model (like hidden size / number of layers) were \emph{manually chosen} with model capacity in mind. For our final experiments with this pretrained model we did not perform any hyperparameter search during training, primarily due to the expensive setup required to train the system. During inference, we tuned the nucleus sampling value from 0.0 to 1.0 in increments of 0.1, choosing the value with the best validation set performance. Our hyperparameter choices for contrastive learning on the retriever have been justified in an ablation study in \appendixref{appendix:ablation_batch_size}. Notably, we use very large minibatches of 12,288 to scale the number of negative examples. To train this model, we used the standard trick of data parallelism across 64 hardware accelerators. This resulted in an effective mini-batch size of 192 per chip, which is small enough to fit a BERT-base sized model on a TPU v3 chip's memory. To accumulate information across different chips before the final softmax, we used the \texttt{tf.tpu.cross\_replica\_sum} function (using an open-source wrapper found \href{https://github.com/google-research/language/blob/master/language/common/utils/tpu_utils.py#L83}{here}).

\subsection{Ablation Study of \retriever}
\label{appendix:ablation_batch_size}

One of our contributions is scaling up a distantly supervised objective for training retrievers on ELI5, originally described in~\citet{jernite2020}. This method uses in-batch negative sampling, making minibatch size a critical hyperparameter for better constrastive learning. We perform controlled experiments initializing our retrievers with REALM-CCNews~\citep{guu2020realm} and varying batch size and keeping all other hyperparameters consistent. In \tableref{tab:crealm_batch_size}, we notice a steady increase in performance as minibatch size is increased, with the largest gains coming by doubling the batch size in~\citet{jernite2020} from 512 to 1024. Finally, in preliminary experiments we saw no benefit of more intelligent negative sampling schemes.

\begin{table}[h]
\begin{center}
\begin{tabular}{ lrr } 
 \toprule
Batch size & R-Prec & Recall@5 \\
\midrule
REALM (pretrained) & 6.6 & 14.9 \\
\midrule
256 & 6.2 & 11.0 \\
512~\citep{jernite2020} & 6.8 & 12.6 \\
1024 & 11.5 & 21.0 \\
12288 (Ours) & 13.3 & 21.2 \\
\bottomrule
\end{tabular}
\end{center}
\caption{The effect of minibatch size on the validation performance of \retriever~. As a baseline, we also add the retrieval performance of the REALM pretrained model which is used as an initialization.}
\label{tab:crealm_batch_size}
\end{table}

Next, we investigate the effect of initialization on the training of ~\retriever~. Unlike~\citet{jernite2020} who initialize their model with BERT, before training we initialize our retriever with a pretrained self-supervised retriever. As a baseline, we initialize our model with ICT, a weaker self-supervised retriever introduced in~\citet{lee-etal-2019-latent}. Both models are trained with minibatch sizes of 12228. In \tableref{tab:crealm_init}, we notice a large improvement in performance when using a better initialization, confirming our design decisions.

\begin{table}[h]
\begin{center}
\begin{tabular}{ lrr } 
 \toprule
Initialization & R-Prec. & R@5 \\
\midrule
REALM (pretrained) & 6.6 & 14.9 \\
\midrule
ICT~\citep{lee-etal-2019-latent} & 9.3 & 16.5 \\
REALM~\citep{guu2020realm} & 13.3 & 21.2 \\
\bottomrule
\end{tabular}
\end{center}
\caption{The effect of initialization on \retriever~. As a baseline, we also add the retrieval performance of the REALM-CCNews pretrained model without any finetuning on ELI5. }
\label{tab:crealm_init}
\end{table}

\subsection{Number of trainable parameters}
\label{appendix:number_parameters}

In \tableref{tab:number_parameters} we present the number of trainable parameters in our model compared to baselines on the leaderboard. Our generator is slightly larger than the models used in prior work, but we utilize a smaller retriever due to the shared query and candidate encoders in REALM. Overall, our system has a similar total number of parameters as baseline models like RAG and BART + DPR.

\begin{table}[h]
\small
\begin{center}
\begin{tabular}{ lrrr } 
 \toprule
 Model & Generator & Retriever & Index \\
 \midrule
 T5-base & 220M & - & - \\
 BART & 406M & - & - \\
 RAG & 406M & 220M & 15B \\
 BART + DPR & 406M & 220M & 15B \\
 RT + \retriever~ & 486M & 110M & 15B \\
\bottomrule
\end{tabular}
\end{center}
\caption{The number of parameters used by our model and baselines. Our generator is slightly bigger than other submissions on the leaderboard, but we use a smaller retriever with a similar sized index.}
\label{tab:number_parameters}
\end{table}

\subsection{Generations from our System}
\label{appendix:generations}

More generations have been provided (along with retrievals, highlighted to show $n$-gram overlap) in the supplementary material (data) as HTML files. We also present a few samples in \tableref{tab:generations_extra}.

\subsection{Human Evaluation Setup}
\label{appendix:human_evaluation}

We conducted several A/B tests between variants of our model using human annotators. We asked a total of 20 participants for help who voluntarily agreed to help with the annotation process. Most participants were English-speaking graduate students in computer science. In every test, participants were shown a question along with two answers (generated by different systems) presented in a random order. They were then asked to choose which generation (1) answered the question better / which answer was more relevant to the question; (2) was more coherent / had less repetition; (3) was more factually correct. Since some annotators had a limited time, we asked them to prioritize question (1) over (2) / (3). Annotators were allowed to select ``Tie'' if they could not choose between the systems. We also permitted them to use search engines, but suggested restricting search to Wikipedia. We present all our results in \tableref{tab:human_evaluation_all}. We also interviewed some participants after the annotation process and discuss our findings in \sectionref{sec:human_eval_difficulty}. Note that while these A/B tests help us understand which system is relatively better, they do not provide an absolute measure of performance~\citep{celikyilmaz2020evaluation} --- annotators reported that there were cases where both answers were very good and other cases where both were very poor. This is a limitation of A/B testing.

\subsection{Effect of length on ROUGE-L}
\label{appendix:length_effect}

In this section we measure the effect of outputs lengths on ROUGE-L scores. To conduct this experiment, we truncate generations by our system to a fixed fraction of tokens across all instances. As we see in \tableref{tab:rouge_length} in the \emph{Truncate} column, shorter generations tend have lower ROUGE-L. To disentangle the effects of length and content, we also measure the generation quality by repeating the truncated generations several times until it matches the original generation length. In the \emph{Repeat $1/f$ times} column, we notice a gap between our model's original generation (24.4 ROUGE-L) and the equal-length truncated generations with repetition. These results indicate that while length helps improve ROUGE-L scores, simple repetition is insufficient.

\begin{table}[h]
\small
\begin{center}
\begin{tabular}{ lrrr } 
 \toprule
 Fraction $f$ & \# Tokens  & Truncate & Repeat $1/f$ times \\
 \midrule
 0.1 & 18.2 & 17.4 & 18.2 \\
 0.2 & 37.0 & 20.8 & 21.1 \\
 0.3 & 55.7 & 22.2 & 22.4 \\
 0.4 & 74.4 & 22.9 & 23.1 \\
 0.5 & 93.4 & 23.4 & 23.6 \\
 0.6 & 112.0 & 23.9 & 23.9 \\
 0.8 & 149.4 & 24.2 & 24.3 \\
 \midrule
 1.0 & 187.3 & 24.4 & 24.4 \\
\bottomrule
\end{tabular}
\end{center}
\caption{Effect of truncating generations (Truncate) from the $p=0.6$ model to keep the first $f$ fraction of tokens, and then repeating the truncated generations $1/f$ times to match the original length (Repeat ...). Notice a consistent increase in ROUGE-L with longer outputs, but a gap between the original generations (24.4) and equal-length generations formed by repeating truncations (Repeat $1/f$ times column).}
\label{tab:rouge_length}
\end{table}

\subsection{More experiments on measuring retrieval grounding of generations}
\label{appendix:grounding_correct_retrievals}

In this section we provide some more experiments testing the grounding of generations in retrieved documents.  Overall, trends are consistent with our observations in \sectionref{sec:grounding}.\\

\begin{figure}[t]
    \centering
    \includegraphics[scale=0.5]{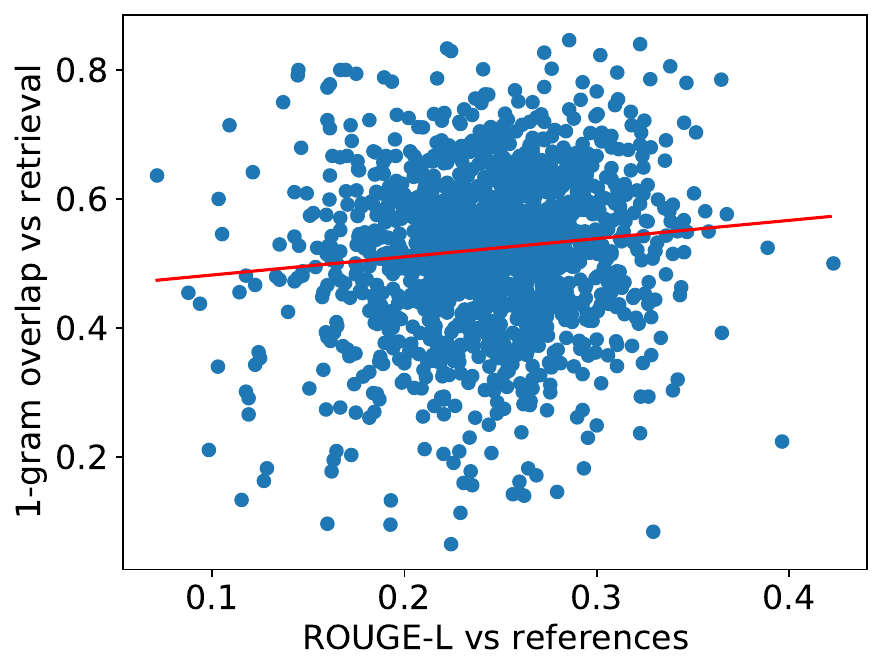}

    \caption{Scatter plot for generations from the $p=0.6$ model between generative quality (ROUGE-L vs reference on X-axis) and grounding with retrieval (unigram overlap with retrieved documents on Y-axis). The plot shows no correlation between the two quantities.}
    \label{fig:scatter_60}
\end{figure}

\noindent \textbf{Scatter plots between generation quality and unigram overlap with retrievals}: We present this scatter plot in \figureref{fig:scatter_60}. There is virtually no correlation between the two quantities, with Spearman $\rho = 0.09$.\\

\begin{figure}[t]
    \centering
    \includegraphics[scale=0.5]{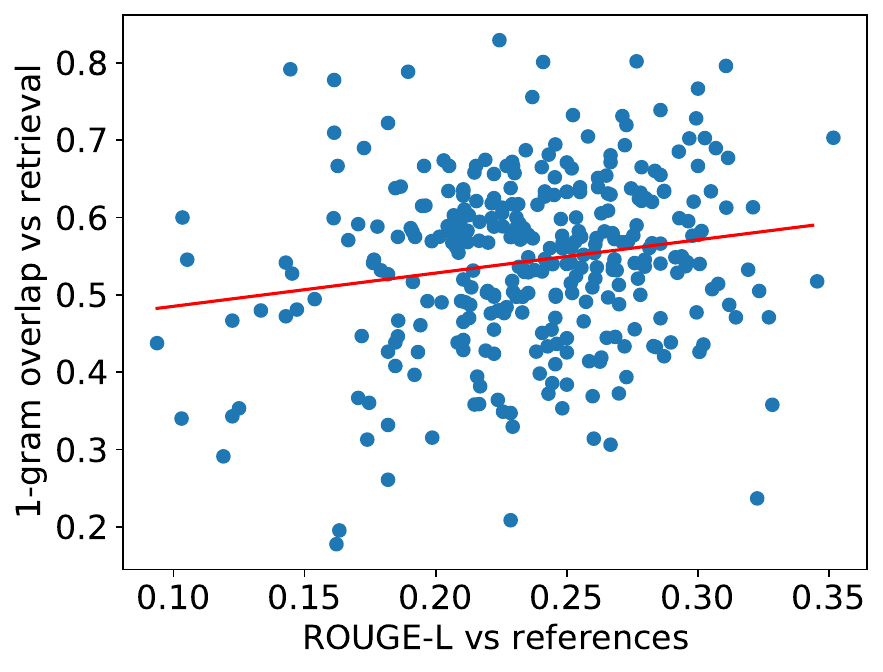}

    \caption{Scatter plot for generations from the $p=0.6$ model between generative quality (ROUGE-L vs reference on X-axis) and grounding with retrieval (unigram overlap with retrieved documents on Y-axis). Unlike \figureref{fig:scatter_60}, this plot only considers \textbf{those cases where ~\retriever~ predicted the correct retrieval}. The plot shows very little correlation between the two quantities (Spearman $\rho = 0.13$).}
    \label{fig:scatter_60_correct_retrieval}
\end{figure}

\noindent \textbf{Instances with correct predicted retrieval}: In \tableref{tab:pred_vs_random_correct_ret}, we present results similar to \sectionref{sec:grounding} considering only those instances where at least one retrieved document matched the gold annotation (roughly 23\% instances). We also present a scatter plot on the same set of instances in \figureref{fig:scatter_60_correct_retrieval} and note a low correlation of $\rho = 0.13$.\\

\begin{table}[h]
\small
\begin{center}
\begin{tabular}{ lrrrrr } 
 \toprule
 & & \multicolumn{2}{c}{vs predicted retr.} & \multicolumn{2}{c}{vs random retr.} \\
 & R-L & ~~~~~~~~~1-g & 2-g & ~~~~~~~1-g & 2-g \\
\midrule
\multicolumn{6}{l}{$p = 0.6$, correct retrieval examples} \\\\
\emph{Predicted} & 23.74 & 54.4 & 10.0 & 39.7 & 4.3  \\
\emph{Random} & 23.91 & 52.5 & 9.6 & 38.8 & 4.0 \\
\midrule
\multicolumn{6}{l}{$p = 0.9$, correct retrieval examples} \\\\
\emph{Predicted}  & 22.40 & 54.9 & 9.2 & 40.9 & 4.3 \\
\emph{Random} & 22.22 & 54.7 & 9.2 & 41.1 & 4.2 \\
\bottomrule
\end{tabular}
\end{center}
\caption{Comparison of generations conditioned on retrievals from ~\retriever~ (Predicted) and randomly chosen retrievals (Random), for \textbf{those cases where ~\retriever~ predicted the correct retrieval}. Notice very small differences in generation quality (R-L) as well as the fraction of $n$-grams ($n$-g) in the generation overlapping with retrievals predicted by \retriever~ (vs predicted retr.). To control for overlap due to stopwords, we also add $n$-gram overlaps with the randomly sampled retrievals.}
\label{tab:pred_vs_random_correct_ret}
\end{table}

\noindent \textbf{Experiments with $p=0.9$}: We conduct additional experiments studying our model variant with higher nucleus sampling values. As we saw in \sectionref{sec:experiments}, these generations tend to be more fluent and coherent, but less relevant to the question. In \tableref{tab:pred_vs_random_all_p90} and \tableref{tab:pred_vs_random_all_nouns_p90} we find consistent trends as \sectionref{sec:grounding}, with very little difference between models conditioned on retrievals from \retriever~ and random retrievals.

\begin{table}[t]
\small
\begin{center}
\begin{tabular}{ lrrrrrr } 
 \toprule
 & & \multicolumn{2}{r}{vs predicted retr.} & \multicolumn{2}{c}{vs random retr.} \\
  & R-L & ~~~~~~~~~1-g & 2-g & ~~~~~~~1-g & 2-g \\
\midrule
\emph{Predicted} & 22.62 & 53.9 & 8.7 & 40.7 & 4.1 \\
\emph{Random} & 22.56 & 53.1 & 8.4 & 40.7 & 4.1 \\
\midrule
\emph{Gold Ans} & - & 54.1 & 9.1 & 40.2 & 3.8 \\
\bottomrule
\end{tabular}
\end{center}
\caption{Comparison of generations (with $p=0.9$) conditioned on retrievals from~\retriever~ (\emph{Predicted}) and randomly chosen retrievals (\emph{Random}). Notice very small differences in: (1) ROUGE-L vs gold answers (R-L); (2) $n$-gram overlap ($n$-g) with retrievals predicted by \retriever~ (vs predicted retr.). \emph{Gold answers} also have a similar overlap with predicted retrievals. To control for overlap due to stopwords, we also add $n$-gram overlaps with the randomly sampled retrievals.}
\label{tab:pred_vs_random_all_p90}
\end{table}

\begin{table}[t]
\small
\begin{center}
\begin{tabular}{ lrrr } 
 \toprule
  & vs qn. & vs predicted retr. & vs random retr. \\
 & & but not in qn. & but not in qn. \\
\midrule
\multicolumn{4}{l}{(lemmatized nouns, proper nouns, numbers only)} \\\\
\emph{Predicted} & 9.1\% & 32.4\% & 12.0\% \\
\emph{Random} & 9.4\% & 30.2\% & 12.3\% \\
\midrule
\emph{Gold Ans} & 8.3\% & 28.8\% & 15.1\% \\
\bottomrule
\end{tabular}
\end{center}
\caption{A fine-grained version of \tableref{tab:pred_vs_random_all_p90} measuring the unigram overlap of nouns/numbers in the generations with the input question (vs qn.), retrievals predicted by \retriever~ (vs predicted retr.) and randomly sampled retrievals (vs random retr.). Similar to \tableref{tab:pred_vs_random_all_p90}, notice very little difference with and without retrieval.}
\label{tab:pred_vs_random_all_nouns_p90}
\end{table}

\begin{table*}[t]
\small
\begin{center}
\begin{tabular}{ lllrrr } 
 \toprule
A & B & Question &  Prefer A & Prefer B & Tie \\
\midrule
\multicolumn{6}{l}{\textbf{Experiment 1}: A comparison between nucleus sampling $p$ values (0.6, 0.9), conditioning on predicted retrievals (pred.).} \\
\multicolumn{6}{l}{\textbf{Result}: Lower entropy more relevant to question, but higher entropy more coherent and lesser repetition.}\\\\
 $p=0.6$, pred. & $p=0.9$, pred. & Which generation answers the question better? & \textbf{41}\% (65) & 30\% (48) & 29\% (46) \\
 & & Which answer is more coherent? & 27\% (42) & \textbf{50}\% (79) & 23\% (37) \\
& & Which ans. is more factually correct + sensical? & 30\% (47) & 37\% (58) & 33\% (52)\\
\midrule
\multicolumn{6}{l}{\textbf{Experiment 2}: A comparison between generations conditioned on predicted (pred.) and random retrievals (rand.).} \\
\multicolumn{6}{l}{\textbf{Result}: Little difference in generation quality / coherence / relevance to question, high amounts of tie.}\\\\
 $p=0.6$, pred. & $p=0.6$, rand. & Which generation answers the question better? & 40\% (78) & 33\% (64) & 27\% (51) \\
  & & Which answer is more coherent?** & 55\% (12) & 27\% (~~6) & 18\% (~~4) \\
  & & Which ans. is more factually correct...** & 48\% (10) & 9\% (~~2) & 43\% (~~9) \\\\
 $p=0.9$, pred. & $p=0.9$, rand. & Which generation answers the question better? & 31\% (52) & 37\% (63) & 32\% (54)\\
  & & Which answer is more coherent? & 32\% (26) & 36\% (30) & 32\% (26) \\
  & & Which ans. is more factually correct + sensical? & 28\% (23) & 35\% (29) & 37\% (30) \\
 \midrule
 \multicolumn{6}{l}{\textbf{Experiment 3}: A comparison between generations conditioned on predicted retrievals (pred.) and the longest gold answer.} \\
 \multicolumn{6}{l}{\textbf{Result}: Strong preference for gold answers over generations.} \\\\
  $p=0.6$, pred. & gold answer & Which generation answers the question better? & 14\% (29) & \textbf{68}\% (138) & 18\% (36) \\
   & & Which answer is more coherent? & 7\% (~~8) & \textbf{71}\% (~~77) & 21\% (23) \\
      & & Which ans. is more factually correct + sensical? & 2\% (~~2) & \textbf{76}\% (~~65) & 22\% (19) \\\\
  $p=0.9$, pred. & gold answer & Which generation answers the question better? & 17\% (49) & \textbf{72}\% (203) & 11\% (31) \\
  & & Which answer is more coherent? & 13\% (14) & \textbf{61}\% (~~65) & 25\% (27) \\
   & & Which ans. is more factually correct + sensical? & 6\% (~~6) & \textbf{72}\% (~~78) & 22\% (24) \\
\bottomrule
\end{tabular}
\end{center}
\caption{Human evaluations experiments with exact number of ratings shown in ($\cdot$). Differences greater than 10\% with more than 50 total samples have been bold marked. The experiments marked with ** have less than 50 samples, so it is difficult to draw meaningful conclusions.}
\label{tab:human_evaluation_all}
\end{table*}

\begin{table*}[t]
    \small
    \centering
    \begin{tabular}{p{2.0\columnwidth}}
    \toprule
    Q: What causes the trail behind jets at high altitude?\\
  \midrule
  A: The same thing that causes the clouds you see at the start of a storm to form in the first place. The air pressure changes as the jet moves through the air, causing it to cool down and condense into water droplets. These droplets are in fact a lot bigger than the water droplets in the clouds, so when the plane passes through them, they can cause the air pressure to change, causing the water droplets to condense into clouds. This is what you see when you look at a cloud formation from the ground. In addition to this, when the plane passes through the clouds, it also moves through the air, causing it to change its temperature, causing the water droplets to condense and form clouds. These clouds are often quite large, and can block out the sun. This is what you see when you look at a contrail. Finally, the water vapor that forms at high altitudes is cooled by the air above, and as it cools, it condenses into clouds. This is what you see when you look at a cumulonimbus cloud. These clouds are even larger than the water droplets in the clouds, and they block out the sun.\\
  \midrule
  Q: How do shampoo and conditioner work?\\
  \midrule
  A: The key ingredients in shampoo and conditioner are usually surfactants, they tend to attract oil from your skin. Shampoo will usually contain an anti-dandruff agent (dandruff is caused by fungi in your skin which cause dandruff). They also contain some moisturizing agents (moisturizers are things that stay in your skin longer and protect your scalp from cleaning), and a lot of anti-dandruff (dandruff is a side effect of high doses of both surfactants and moisturizers) Conditioner will contain moisturizing agents, like a skin moisturizer. They tend to cause oils to be separated from your hair (e.g. wash your hair with conditioner instead of shampoo). They both have a foaming agent that makes them foam up (usually Hydrogen Peroxide). They also have a foaming agent that stops dirt from attaching to your hair. Some conditioner will also suppress the action of the shampoo (to help prevent dirt from attaching to your hair). So what's the difference between shampoo and conditioner? Well, if you combine both the shampoo and conditioner\\
  \midrule
  Q: When glancing at a clock, why does the first second after glancing at it sometimes feel longer than the rest?\\
  \midrule
  A: This is known as chronostasis. Your brain gets used to seeing the second hand move after the first, so it ignores the first one. If you were to look at the clock a lot, you'd see the second hand move a lot more than the first, and you'd be able to tell the difference between the second and the first. The second hand will appear to move much slower than the first hand because your brain is used to seeing the second hand move a lot more than the first. If you were to go on a long trip, you'd get used to the second hand moving a lot more than the first, and you'd have no problem telling the difference between the second and the first hand. It's kind of like how you don't notice your nose when you're driving, but if you look at it for a while, you start to notice it. Your brain is just getting used to seeing the nose move a lot more than the other, and it can't tell the difference between the two. It's kind of like that with your eyes. \\
  \midrule
  Q: Why do Britain and other English empire countries still bow to monarchs? What real purpose does the queen serve?\\
  \midrule
  A: Depends what you define as "purpose". There is no specific purpose for the British monarch - rather, they are figureheads that serve a role. The role is to have a high degree of influence over a country, though the rules of the monarchy can somewhat vary from country to country, and the ceremony of "kissing the royal corpse" does not always involve kissing the monarch. Whether that's the only reason for the custom, or if it was the only reason, I can't say, but that's the reasoning. When the U.K. was in full power, the monarch was a puppet of Parliament, and the powers of the monarch were transferred to the Prime Minister and thus to the Prime Minister's deputy, who then became the Prime Minister. Since then, Parliament has been able to vote on legislation that goes through the monarch, although they may still act as the monarch's representative in negotiating treaties, which can have very very deep consequences. The Queen's role, as a representative of Britain, doesn't necessarily involve her formal approval of any of the laws or legislation that goes through Parliament, though. All that's left is for the legislature to pass laws \\
  \midrule
  Q: What exactly is fire, in detail? How can light and heat come from something we can't really touch? \\
  \midrule
  A: Fire is the chemical reaction of fuel (oxygen) with (what we call) impurities (ash, soot, oil etc). These impurities are created due to a number of things. If you were to study the chemical composition of a log (or a candle) and analyse it you would discover the chemical formula (A, C, and G) which is a simplified version of what actually happens. There is always fuel, and there is always the ash/soot. Wood is made up of mostly carbon and hydrogen and these are the fuels that create the fire. Hydrogen and oxygen are reactive, and they form water (H2O) and CO2 (CO2). If we split a log into smaller pieces, we start to see the process of combustion as it occurs. The carbon is burned, and with the explosion of CO, H2O and H2CO2, we get water and carbon dioxide (mostly). If we smash a log together, we start to see what happens when we also hit another log with another log...imagine smashing two sand castles together. We see how the sand castles and brick castles form and collapse at different rates...then it starts to burn. The smoke, ash and flames are the actual combustion of fuel. \\
  \midrule
  Q: What goes on in those tall tower buildings owned by major banks? \\
  \midrule
  A: The actual buildings are not there for profit but for show. There are many real world historical buildings that have little to no effect other than being a symbol of an economic entity or symbol of an era or location, and nothing more. For example look at Sears, Sears didn't care what went on inside, it was all about the \_appearance\_ of its location, the prestige of the location, the facilities and so on. It didn't care about how long it took it to operate, it was about how much people would pay to go see it. Sears was a landmark as a cultural movement and other big companies followed suit, so if you want to see a building you've never seen before, you have to go see Sears, just like you have to see a Toyota Camry for Toyota Camry. They used to be all about building new factories, some of them if I recall, but now that they're bigger, that means that more factory jobs are coming to them. You've probably seen them in stores as stores where people buy and sell stuff, so there aren't that many places for them to come from. Instead, it's just for show, a symbol of rich people.  \\
  \bottomrule
       \end{tabular}
    \caption{Example generations from our LFQA system with \(p=0.9\).}
    \label{tab:generations_extra}
\end{table*}

\end{document}